\ifdefined\XeTeXversion\PassOptionsToPackage{xetex}{hyperref}\else\pdfoutput=1\fi

\documentclass[11pt]{article}

\usepackage[final]{acl}

\usepackage{times}
\usepackage{latexsym}
\usepackage[T1]{fontenc}
\usepackage[utf8]{inputenc}
\usepackage{microtype}
\usepackage{inconsolata}
\usepackage{graphicx}
\graphicspath{{./}{../}}
\usepackage{hyperref}
\usepackage{url}
\usepackage{booktabs}
\usepackage{xcolor}
\usepackage{colortbl}
\usepackage{multirow}
\usepackage{siunitx}
\usepackage{subcaption}
\usepackage{amsmath}
\usepackage{amssymb}
\usepackage{amsfonts}
\usepackage{float}
\usepackage{placeins}
\usepackage{array}
\newcolumntype{L}[1]{>{\raggedright\arraybackslash}p{#1}}
\newcolumntype{C}[1]{>{\centering\arraybackslash}p{#1}}
\newcolumntype{R}[1]{>{\raggedleft\arraybackslash}p{#1}}

\title{Reference-Based Analysis of Coherence and Diversity in \\Open-Ended Text Generation}

\author{Esteban Garcés Arias \\
  Department of Statistics, LMU Munich \\
  Munich Center for Machine Learning (MCML) \\
  \texttt{Esteban.GarcesArias@stat.uni-muenchen.de} \\}

\begin{document}
\maketitle

\begin{abstract}
Evaluating open-ended text generation involves understanding how different properties of a continuation relate to its perceived quality. We present a reference-based framework for examining coherence and diversity through three perspectives: aligning their evolution with human trajectories, comparing their summaries with a human continuation of the same prompt, and estimating their likelihood under a human reference distribution. Experiments with human quality ratings suggest that diversity-based alignment and mean-based comparisons capture quality-related variation, although the comparisons do not establish a predictive advantage for temporal alignment over simpler baselines. Reference likelihood also shows positive associations with ratings, with results varying across reference configurations and scoring horizons. Together, these analyses provide a structured way to examine how measured coherence and diversity relate to human judgments, while distinguishing similarity to human references from quality itself. Code and analysis resources are available at \url{https://github.com/EstebanGarces/likely_human}.
\end{abstract}

\section{Introduction}
Evaluating open-ended text generation is difficult because many different continuations can be appropriate for the same prompt. Reference overlap alone cannot express the full range of valid outputs, and quality includes several dimensions that can conflict \citep{gatt2018survey,celikyilmaz2020evaluation,howcroft2020twenty}. A continuation may be fluent but repetitive, or lexically varied but inconsistent with its context. Coherence and diversity therefore provide useful, though incomplete, perspectives on the quality of generated text \citep{hashimoto2019unifying,su2022contrastive}.

\begin{figure}[t]
\centering
\includegraphics[width=\columnwidth]{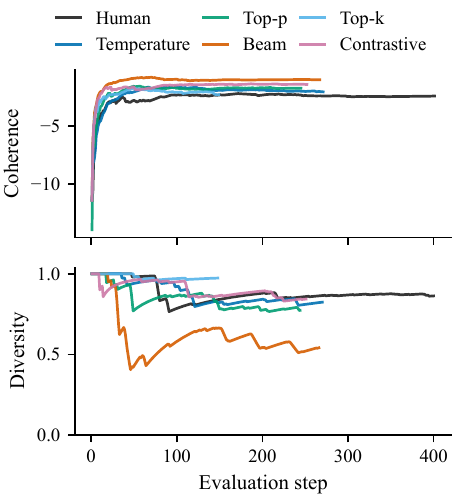}
\caption{How coherence (top) and diversity (bottom) evolve within human and machine continuations of the same prompt. Each curve follows one continuation as more of its text is evaluated. Higher coherence values indicate greater model predictability; higher diversity values indicate less $n$-gram repetition. The framework compares these curves through alignment, summaries, and human reference distributions (Table~\ref{tab:facets}).}
\label{fig:trajectories}
\end{figure}

Figure~\ref{fig:trajectories} illustrates how coherence and diversity measurements evolve as successive prefixes of a continuation are evaluated. These trajectories offer a view of the text beyond its final score. They also raise a broader question: which aspects of the measurements are useful when comparing a generated continuation with human writing?

We examine this question through three analytical facets. \textbf{Temporal alignment} compares the shape of a trajectory with human trajectories. \textbf{Summary comparison} examines how its average values and variability differ from those of a human continuation of the same prompt. \textbf{Reference-distribution likelihood} assesses how typical a joint coherence--diversity profile is within a human corpus. These perspectives share measurements but use human references in different ways, allowing us to examine both sequential patterns and aggregate properties.

We use coherence and diversity as operational measures of model predictability and $n$-gram non-repetition. They capture selected properties relevant to generation, while discourse coherence and semantic diversity remain broader concepts. Our experiments examine how the resulting scores relate to overall-quality ratings of English machine continuations.

Three research questions guide the analysis:
\begin{enumerate}
\item \textbf{RQ1: Predictive information.} How do temporal and prompt-relative summary scores relate to human ratings compared with simpler baselines?
\item \textbf{RQ2: Length and references.} How sensitive are those associations to reference selection, length normalization, and the number of scored tokens?
\item \textbf{RQ3: Reference typicality.} How does joint-profile likelihood relate to ratings, and how does that relationship depend on feature preparation and reference modeling?
\end{enumerate}

We contribute a framework that connects these three uses of human references, an empirical comparison with simple and established baselines, and sensitivity analyses of length and reference design. The experiments combine an analysis of archived full trajectories with controlled measurements at fixed token horizons. Together, they examine which relationships with human judgments persist across these choices and whether temporal comparison offers additional predictive information.

\section{Related Work}
\paragraph{Evaluation of open-ended generation.}
BLEU \citep{papineni2002bleu}, ROUGE \citep{lin2004rouge}, and METEOR \citep{banerjee2005meteor} measure forms of lexical overlap. Their interpretation is limited when a prompt admits many acceptable continuations. Surveys emphasize both the diversity of evaluation criteria and the difficulty of relating automatic metrics to human judgments \citep{novikova2017why,sai2022survey,howcroft2020twenty}. Our study focuses on the relationship between reference-based scores and overall-quality ratings of individual continuations.

\paragraph{Coherence and diversity.}
The tension between repetition and unpredictable output motivates contrastive search \citep{su2022contrastive}, top-$k$ sampling \citep{fan2018hierarchical}, and nucleus sampling \citep{holtzman2019curious}. Comparative work shows that decoding preferences depend on the setting \citep{garces-arias-etal-2025-decoding}. Adaptive Contrastive Search \citep{garces-arias-etal-2024-adaptive} and GUARD \citep{ding-etal-2025-guard} use entropy signals over generation steps to adapt decoding. Min-$k$ Sampling uses local changes in sorted logits to determine truncation boundaries independently of temperature scaling \citep{ding-etal-2026-min}. Variance-Calibrated Modulation reshapes the distribution before truncation through contextual token weighting and a logit-scale-aware repetition penalty \citep{ding2026breakinglikelihoodtrapvariancecalibrated}. These methods modify generation, complementing our analysis of coherence and diversity in the resulting continuations.

\paragraph{Composite and distributional metrics.}
BERTScore \citep{zhang2020bertscore} and BARTScore \citep{yuan2021bartscore} use contextual representations or generation probabilities. MAUVE \citep{pillutla2021mauve} compares distributions of human and machine text. Q*Text \citep{arias2025betteropenendedtextgeneration} combines normalized coherence, diversity, and perplexity through weighted Gaussian penalties. We distinguish its fixed initialization from parameters optimized against human ratings. We report MAUVE at the corpus level and evaluate the other scores through continuation-level correlations, reflecting their different units of analysis. Discourse-oriented measures such as DiscoScore \citep{zhao2022discoscore}, prompted evaluators such as G-Eval \citep{liu-etal-2023-g}, and controlled diagnostic benchmarks such as OpenMEVA \citep{guan-etal-2021-openmeva} address complementary questions.

\paragraph{Human-reference comparisons.}
Generalized stochastic dominance compares decoding strategies across criteria and measurement scales \citep{garces-arias-etal-2025-statistical}; we examine continuation-level score correlations. Credal-set analysis compares human and model variation across multiple continuations \citep{garces-arias-etal-2025-geometry}, whereas our diversity signal measures repetition within a continuation. Work on token exclusion under decoding distinguishes detectability from perceived quality \citep{garcesarias2026truncation}. Its quality analysis shares the rating collection used here and therefore provides related context rather than independent validation.

\paragraph{Temporal alignment.}
DTW aligns numerical sequences that evolve at different rates \citep{sakoe1978dynamic,muller2007dynamic}. Applied here, DTW aligns numerical coherence and diversity trajectories. The resulting paths locate differences in those measurements; interpreting them in terms of entities or narrative events would require additional linguistic analysis. Soft-DTW \citep{cuturi2017soft} and explicit slope or change-point descriptors offer alternative temporal analyses; we use the DTW configuration specified in Section~\ref{sec:temporal}.

\section{Data and Measurement}
\label{sec:data}
Table~\ref{tab:setup} summarizes the texts, human ratings, and measurement settings. We first analyze the available full trajectories, then use a shared measurement procedure on fixed-length prefixes to examine sensitivity to length and extraction settings.
\begin{table*}[t]
\centering\small
\begin{tabular}{@{}p{0.22\textwidth}p{0.74\textwidth}@{}}
\toprule
Component & Design \\
\midrule
Generation & Reported generator: Qwen2.5-7B. Temperature $\tau=0.9$; top-$p$, $p=0.95$; top-$k$, $k=50$; beam search, $B=5$; contrastive search, $k=10$, $\alpha=0.6$. \\
Primary collection & 50 prompts (25 Wikinews, 25 WikiText), each with one human and five machine continuations. Correlations evaluate the 250 machine texts; 50 human texts supply references. \\
Human evaluation & Two five-point overall-quality ratings for each of the 300 texts; their arithmetic mean is the primary target. \\
Extended references & 10,000 human records from BookCorpus, ROCStories, Wikinews, WikiText, and WritingPrompts; 9,877 retained for archived reference fitting. \\
Archived analysis & Released full-trajectory features; incomplete extraction provenance. Temporal alignment, prompt-relative summaries, and reference-profile likelihood. \\
Controlled analysis & Shared OPT-2.7B extraction at fixed continuation horizons. At 64 tokens: 40 prompt groups, 200 machine texts, 7,686 corpus references. At 128 tokens: 23 groups, 115 machine texts, 7,149 corpus references. \\
Comparisons & Endpoint and trajectory summaries, PCA, and fixed Q*Text; MAUVE supplies separate corpus-level context. Spearman correlations with prompt-cluster bootstrap intervals. \\
\bottomrule
\end{tabular}
\caption{Overview of the datasets, generation configurations, and evaluation procedures. Fixed-horizon analyses retain complete prompt groups and exclude human corpus references that exactly overlap with the primary human continuations.}
\label{tab:setup}
\end{table*}

\subsection{Prompts, Continuations, and Ratings}
A \emph{prompt} is the fixed input text $x$. A \emph{continuation} is the complete subsequent text $y=(y_1,\ldots,y_L)$, written by a person or a model. A \emph{continuation prefix}, $y_{1:t}$, is only its first $t$ evaluation tokens. A trajectory records measurements on successive prefixes of that same continuation. Each record therefore links the prompt and continuation to an ordered array of measurements.

The continuations and human ratings come from prior work \citep{garces-arias-etal-2025-decoding,garces-arias-etal-2025-statistical}; Table~\ref{tab:setup} records their settings. The prompts are expository, so the evaluation concerns open-ended continuation in that setting. Human texts supply references and are excluded from the main score--rating correlations. Prompt groups remain intact in uncertainty estimation.

Agreement between the annotators is linearly weighted Cohen's $\kappa=0.324$ and Spearman's $\rho=0.529$, motivating per-rater analysis and caution about fine-grained rankings. Continuation lengths vary substantially across methods; Appendix~\ref{app:data} reports their ranges and further agreement statistics.

\subsection{The Extended Human Corpus}
\label{sec:corpus}
The extended collection supplies human references from five domains (Table~\ref{tab:setup}). We retain its stored prompt and continuation boundaries. Missing cached features and exact overlap with primary human continuations reduce the archived reference set to 9,877 records. These exclusions do not rule out near duplicates or shared documents. Appendix~\ref{app:data} details segmentation and exclusions; controlled-horizon eligibility is documented in Appendix~\ref{app:length}.

\subsection{Quality Trajectories and Cache Preparation}
\label{sec:features}
Equation~\ref{eq:coherence} defines the intended coherence signal as a running average of token log probabilities under the external model OPT-2.7B \citep{zhang2022optopenpretrainedtransformer}:
\begin{equation}
c_t=\frac{1}{t}\sum_{i=1}^{t}\log p_{\mathcal M}(y_i\mid x,y_{<i}).
\label{eq:coherence}
\end{equation}
Higher values indicate that the continuation is more predictable under the external model. We use this as a coherence proxy, recognizing that discourse-level properties such as logical consistency and entity continuity extend beyond predictability. The full-trajectory analysis uses released numerical trajectories. The controlled experiment computes token-aligned coherence and diversity measurements over prefixes of all 300 primary and 10,000 reference texts (Section~\ref{sec:length}). The primary cache stores running coherence and diversity scores but lacks a complete extraction configuration. The controlled experiment implements Equations~\ref{eq:coherence} and~\ref{eq:diversity}; the conditioning, tokenization, and exact implementation underlying the archived values remain unverified. This limits interpretation of the archived scores even where their downstream calculations can be reproduced.

Let $G_n(y_{1:t})$ be the multiset of contiguous token $n$-grams, including repeated occurrences. Equation~\ref{eq:diversity} defines the intended diversity signal as the product of non-repetition ratios for sizes two through four:
\begin{equation}
d_t=\prod_{n=2}^{4}\frac{|\operatorname{unique}(G_n(y_{1:t}))|}{|G_n(y_{1:t})|}.
\label{eq:diversity}
\end{equation}
An empty $n$-gram set contributes one. Values near one indicate little repeated $n$-gram material, giving a lexical measure of diversity. Adding unique material can increase a ratio after an earlier decrease, so the trajectory need not be monotone. For archived trajectories, the horizontal coordinate follows the stored evaluation steps.

The primary coherence cache contains $L$ values per text and diversity contains $L-1$; we calculate summaries over the available values without padding. In the extended cache, we first convert token log probabilities into cumulative means so that the coherence summaries describe the same intended quantity. The underlying arrays are rounded and their extraction configurations are incomplete. We therefore treat the cached likelihood analysis as exploratory and evaluate extraction-compatible prefixes separately. Appendix~\ref{app:audit} records the preparation steps and remaining uncertainty.

\section{Three Analytical Facets}
Table~\ref{tab:facets} summarizes how each facet uses the measurements and human references. Temporal alignment follows the progression of the measured signals through a continuation; summary comparison and likelihood examine their aggregate profiles. All three derive from cumulative prefix measurements and can therefore be affected by token order, although only temporal alignment explicitly matches positions across trajectories.
\begin{table*}[t]
\centering\small
\setlength{\tabcolsep}{3pt}
\begin{tabular}{@{}p{0.19\textwidth}p{0.23\textwidth}p{0.24\textwidth}p{0.28\textwidth}@{}}
\toprule
Facet & Representation & Human reference & Interpretation \\
\midrule
Temporal alignment & Ordered values of each signal & A collection of trajectories; paired reference in an ablation & Similarity of aligned trajectories \\
Summary comparison & Means and variances & Continuation of the same prompt & Signed difference from the matched human \\
Reference likelihood & Joint mean or variance profile & Distribution fitted to corpus profiles & Typicality under that reference model \\
\bottomrule
\end{tabular}
\caption{Three ways to compare coherence and diversity with human references. Temporal alignment compares ordered measurements, summary comparison uses the human continuation of the same prompt, and likelihood describes the profile within a human corpus.}
\label{tab:facets}
\end{table*}

\subsection{Temporal Alignment}
\label{sec:temporal}
The temporal facet compares a machine trajectory $a$ with human trajectories $b$ using DTW. Let $D(a,b)$ denote the alignment cost with scalar absolute differences, closed endpoints, and the \texttt{symmetric2} step pattern. We use no warping window. Appendix~\ref{app:dtw} gives the recurrence and boundary conditions needed to reproduce the scores.

Let $H$ contain the 50 human reference trajectories for the signal being compared. The primary temporal score averages their length-normalized distances:
\begin{equation}
s_{\mathrm{DTW}}(a)=-\frac{1}{50}\sum_{b\in H}
\frac{D(a,b)}{(|a|+|b|)/2}.
\label{eq:dtw}
\end{equation}
We use the negative distance so that larger values indicate closer temporal alignment. 

The all-reference score characterizes resemblance to the observed human trajectory collection; it does not condition reference selection on the evaluated prompt. To distinguish that choice, an ablation uses only the human continuation of the same prompt. Within this paired setting, we also remove length normalization while leaving the references and DTW recurrence fixed. This isolates the effect of distance normalization within the paired-reference comparison.

An alignment path provides a way to inspect where measured trajectories differ. Its usefulness for explaining particular textual properties, such as repetition or topic shifts, remains an empirical question.

\subsection{Summary Comparison}
\label{sec:summary}
For each trajectory we calculate its mean and population variance. Here $L_m$ denotes the machine coherence-array length. Let $\bar c_m,\bar d_m$ be the machine means and $\bar c_h,\bar d_h$ the means of the human continuation for the same prompt. The normalized signed differences are
\begin{equation}
\Delta_c=\frac{(\bar c_m-\bar c_h)}{L_m},\qquad
\Delta_d=\frac{(\bar d_m-\bar d_h)}{L_m}.
\label{eq:differences}
\end{equation}
We standardize the differences in Equation~\ref{eq:differences} over the 250 machine observations, using the sample standard deviation. Equation~\ref{eq:summary} averages the resulting components:
\begin{equation}
s_{\mathrm{mean}}=\tfrac12\{z(\Delta_c)+z(\Delta_d)\}.
\label{eq:summary}
\end{equation}
The variance composite replaces trajectory means with trajectory variances; the overall composite averages all four standardized differences. Their signs are retained as defined, rather than selected to maximize correlation.

The composite expresses a signed contrast with the matched human continuation. Standardization centers each component, allowing positive and negative differences to offset one another. Its value depends on the collection used for standardization; application to new texts would require a fixed normalization reference.

Division by $L_m$ weights a fixed difference more heavily for shorter texts. We examine this length-dependent choice through an ablation that removes the division and refits standardization. For equal-length texts, the division would leave the standardized scores unchanged.

\subsection{Likelihood}
\label{sec:likelihood}
The likelihood facet describes typicality of a joint summary vector $u$. We fit separate bivariate Gaussian reference models to the mean pair $(\bar c,\bar d)$ and variance pair $(v_c,v_d)$. The cached reconstruction fits the sample mean $\mu$ and covariance $\Sigma$ to 9,877 retained human records. The controlled experiment refits these parameters using eligible records at each token horizon. We evaluate the log density in Equation~\ref{eq:likelihood},
\begin{align}
s_{\mathrm{lik}}(u)={}&-\log(2\pi)-\tfrac12\log|\Sigma|\nonumber\\
&-\tfrac12(u-\mu)^T\Sigma^{-1}(u-\mu),
\label{eq:likelihood}
\end{align}
which avoids numerical underflow while preserving the ranking of the density. Fitted parameters are provided in Appendix~\ref{app:likelihood}.

Its relationship to quality is assessed through the human ratings. Diversity is bounded and variances are nonnegative, whereas a Gaussian has unbounded support. We therefore inspect the distributions and compare the full Gaussian with a diagonal-covariance Gaussian and a Gaussian-kernel density estimate using Scott's bandwidth. The alternatives use the same reference records and features, without rating-based tuning. We also fit each reference domain separately to assess sensitivity to corpus composition.

The archived likelihood comparisons remain exploratory because of the extraction limitations in Section~\ref{sec:features}. The controlled experiment evaluates likelihood with consistent extraction across both corpora.

\section{Evaluation Protocol and Baselines}
\label{sec:protocol}
\paragraph{Common evaluation unit.}
The full-trajectory comparisons use the same 250 machine texts and the mean of their two ratings. Fixed-horizon analyses use eligible subsets. We calculate Spearman's rank correlation with average ranks for ties. Full-trajectory scores use the archived measurements described in Section~\ref{sec:features}.

\paragraph{Uncertainty and comparisons.}
We use 3,000 prompt-cluster bootstrap replicates, retaining all five machine outputs and their ratings within each sampled group. Sample-dependent score normalizations are refitted; human references and PCA loadings remain fixed. Intervals therefore condition on the reference resources and do not capture uncertainty from new annotators. Appendix~\ref{app:data} specifies the resampling and normalization details.

We report percentile 95\% intervals for individual correlations and for paired differences calculated within the same replicate. A paired interval containing zero leaves the direction of the difference unresolved; it is not evidence of equivalence. The ablations are exploratory and their intervals are not adjusted for testing multiple alternatives. We additionally report correlations against each annotator to reveal dependence on the rating source.

\paragraph{Experimental comparisons.}
For RQ1, we compare temporal and summary scores with simpler and established baselines. For RQ2, we vary references and normalization, then evaluate fixed token horizons. For RQ3, we examine reference preparation and density models using archived and controlled measurements. All analyses use full-text ratings; Appendices~\ref{app:audit} and~\ref{app:length} give implementation details.

\paragraph{Endpoint and summary baselines.}
The final cached coherence and diversity values provide endpoint baselines. The trajectory mean and variance of each signal help assess whether aggregate properties capture similar associations to temporal comparison. These baselines require no human reference trajectory at scoring time.

\paragraph{PCA and fixed Q*Text.}
PCA provides an unsupervised combination of trajectory means and variances. We retain the released fit on all 300 texts and evaluate its fixed components on the 250 machines; component signs are arbitrary. Fixed Q*Text applies the published, untuned initialization to cached endpoint perplexity, coherence, and diversity \citep{arias2025betteropenendedtextgeneration}. Its normalization and exact formula are given in Appendix~\ref{app:primary}. Neither baseline is fitted to the quality ratings in this comparison.

\paragraph{Corpus-level context.}
We report the original MAUVE results descriptively because they compare whole text distributions, whereas our primary analysis evaluates individual continuations. MAUVE supplies one corpus-level value for each decoding strategy. Appendix~\ref{app:primary} documents their inputs and settings.

\section{Results}
\label{sec:results}
\subsection{RQ1: Temporal and Summary Associations}
\label{sec:rq1}
\begin{table}[t]
\centering\small
\setlength{\tabcolsep}{3pt}
\begin{tabular}{@{}lcc@{}}
\toprule
Score & $\rho$ & 95\% interval \\
\midrule
Fixed Q*Text & $0.612$ & $[0.510,0.677]$ \\
Temporal: diversity & $0.536$ & $[0.442,0.615]$ \\
Temporal: coherence & $0.146$ & $[0.012,0.285]$ \\
Summary: mean & $0.527$ & $[0.399,0.639]$ \\
Summary: variance & $-0.525$ & $[-0.645,-0.367]$ \\
Summary: all four & $-0.106$ & $[-0.273,0.009]$ \\
Endpoint diversity & $0.511$ & $[0.383,0.615]$ \\
Mean diversity & $0.493$ & $[0.372,0.593]$ \\
PCA: PC1 & $-0.376$ & $[-0.479,-0.256]$ \\
PCA: PC2 & $-0.535$ & $[-0.647,-0.402]$ \\
\bottomrule
\end{tabular}
\caption{Association between each score and mean human quality ratings on 250 machine texts. Values are Spearman correlations with 95\% prompt-cluster bootstrap intervals. Temporal scores use negative distances; PCA signs follow the released components.}
\label{tab:primary}
\end{table}

\begin{figure*}[t]
\centering\includegraphics[width=\textwidth]{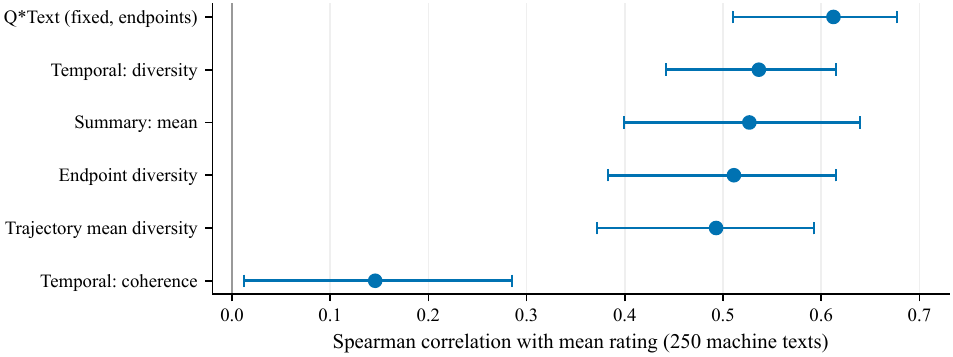}
\caption{Association between each score and mean human quality ratings on the same 250 machine texts. Points show Spearman correlations and bars show 95\% prompt-cluster bootstrap intervals. Temporal scores use negative DTW distances, so higher values indicate closer alignment. Paired comparisons between scores are reported in the text.}
\label{fig:correlations}
\end{figure*}

Diversity alignment shows a stronger association with ratings than coherence alignment in the full-trajectory analysis (Table~\ref{tab:primary}; Figure~\ref{fig:correlations}). Negative diversity DTW correlates with mean quality ratings at $\rho=0.536$ (95\% interval $[0.442,0.615]$), while negative coherence DTW reaches 0.146 ($[0.012,0.285]$). The stronger diversity association may reflect lexical repetition, limitations of the coherence proxy, or differences between decoding methods; the overall-quality ratings do not distinguish those explanations.

The summary mean composite reaches $\rho=0.527$ ($[0.399,0.639]$). Its variance counterpart correlates negatively ($-0.525$), and averaging all four components gives $-0.106$. Combining all components gives a weaker association in this sample, consistent with contributions in different directions (Section~\ref{sec:summary}).

Endpoint diversity alone reaches $\rho=0.511$, and its trajectory mean reaches 0.493. The paired difference between diversity DTW and endpoint diversity is 0.025, with interval $[-0.082,0.141]$. Against the trajectory mean, the difference is 0.044 ($[-0.060,0.145]$). For RQ1, these comparisons leave the predictive advantage of temporal alignment unresolved.

\begin{figure*}[t]
\centering\includegraphics[width=\textwidth]{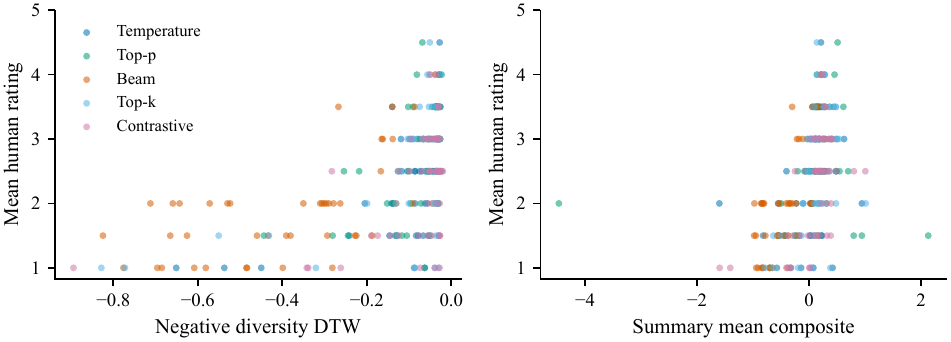}
\caption{How diversity alignment (left) and the mean summary composite (right) relate to human quality ratings. Each point represents a machine continuation, colored by decoding strategy; some points overlap. The spread illustrates variation both within and between strategies.}
\label{fig:scores}
\end{figure*}

The paired difference between diversity DTW and the summary mean composite is 0.010 ($[-0.125,0.145]$). The interval permits meaningful differences in either direction, leaving their relative performance uncertain. Figure~\ref{fig:scores} shows the score--rating relationships by decoding strategy, including substantial within-score disagreement.

\subsection{RQ1: Baselines and Annotator Sensitivity}
Fixed Q*Text reaches $\rho=0.612$ ($[0.510,0.677]$), the largest positive point estimate among the primary scores in Table~\ref{tab:primary}. Its parameters use the published initialization without fitting to these ratings. The paired difference for diversity DTW minus fixed Q*Text is $-0.076$ ($[-0.141,0.021]$), which does not resolve their ordering.

The PCA results in Table~\ref{tab:primary} provide further context: the magnitude of PC2's association is similar to that of diversity DTW. Its arbitrary orientation and unsupervised objective, however, differ from a quality-calibrated score.

Table~\ref{tab:raters} shows that associations also vary by annotator. For example, the summary mean composite correlates 0.554 and 0.300 with the two raters, while diversity DTW correlates 0.507 and 0.436. These differences suggest caution when interpreting agreement with the average rating as a stable preference. Agreement between raters provides context for this variability, although it does not set a numerical ceiling on correlation with their average.

Corpus-level MAUVE ranks temperature sampling first (0.3207), followed by beam search (0.1006), contrastive search (0.0921), top-$k$ (0.0524), and top-$p$ (0.0225). These distributions do not reproduce the order of average human quality ratings: top-$p$ has the highest machine rating, 2.69, and beam search the lowest, 2.00. With 50 texts per distribution, this comparison offers limited descriptive context for how corpus similarity and individual quality judgments may differ.

\subsection{RQ2: Length and Reference Ablations}
\label{sec:rq2}
\begin{table}[t]
\centering\small
\setlength{\tabcolsep}{3pt}
\begin{tabular}{@{}lcc@{}}
\toprule
Score & $\rho$ & 95\% interval \\
\midrule
DTW div., all references & $0.536$ & $[0.442,0.615]$ \\
DTW div., paired & $0.512$ & $[0.415,0.593]$ \\
DTW div., paired, raw & $0.436$ & $[0.326,0.530]$ \\
DTW coh., paired & $0.254$ & $[0.102,0.397]$ \\
DTW coh., paired, raw & $0.181$ & $[0.036,0.321]$ \\
Summary, divided by $L$ & $0.527$ & $[0.399,0.639]$ \\
Summary, no division & $0.482$ & $[0.349,0.585]$ \\
\bottomrule
\end{tabular}
\caption{Length and reference ablations. Paired means one human reference for the same prompt. Raw DTW omits length normalization.}
\label{tab:ablations}
\end{table}

Table~\ref{tab:ablations} addresses RQ2 by varying reference selection and normalization. Using only the human trajectory for the same prompt gives a diversity DTW correlation of 0.512. Removing length normalization from this same paired-reference procedure lowers the observed correlation to 0.436. The paired difference is 0.076 ($[0.044,0.108]$), supporting length normalization for this particular diversity comparison. Its scope is the paired diversity comparison evaluated here.

For coherence, the paired-reference normalized score reaches 0.254, compared with 0.181 without normalization and 0.146 for the all-reference score. The two signals show different patterns across reference choices, suggesting that the usefulness of a matched reference may depend on the measured property.

The unweighted summary mean composite reaches 0.482, compared with 0.527 after division by machine length. The paired difference is 0.044 ($[-0.075,0.183]$), which does not resolve an advantage from that weighting.

\subsection{RQ2: Controlled Token Horizons}
\label{sec:length}
\begin{table*}[t]
\centering\small
\begin{tabular}{lcc}
\toprule
Score & 64 tokens ($n=200$) & 128 tokens ($n=115$)\\
\midrule
DTW coherence & 0.037 [-0.113, 0.199] & 0.177 [0.017, 0.341] \\
DTW diversity & 0.330 [0.222, 0.422] & 0.440 [0.282, 0.571] \\
Gaussian mean & 0.128 [-0.015, 0.276] & 0.296 [0.160, 0.419] \\
Gaussian variance & 0.338 [0.206, 0.456] & 0.571 [0.428, 0.687] \\
Endpoint diversity & 0.388 [0.277, 0.496] & 0.547 [0.417, 0.655] \\
Summary mean & 0.268 [0.101, 0.423] & 0.353 [0.147, 0.536] \\
\bottomrule
\end{tabular}
\caption{Associations with full-text quality ratings when all scored continuations have a common token horizon. Values are Spearman correlations with 95\% prompt-cluster bootstrap intervals. Each column uses the eligible texts and references for that horizon; the paired comparison in Appendix~\ref{app:length} holds membership constant.}
\label{tab:controlled}
\end{table*}

To examine the associations at a common scored length, we measure coherence and diversity using the same OPT-2.7B configuration for both corpora. We evaluate the first 64 and 128 continuation tokens in separate analyses, without padding shorter texts. We retain complete prompt groups and refit the human reference models at each horizon; Table~\ref{tab:setup} reports the eligible collections.

Table~\ref{tab:controlled} reports positive diversity-DTW and variance-profile likelihood associations at both horizons. Adjusting for original continuation and prompt lengths retains positive associations. Separately adjusting score and rating ranks for decoding strategy also retains positive associations for both scores, whereas coherence-DTW intervals include zero at both horizons (Appendix~\ref{app:length}, Table~\ref{tab:strategy}). This suggests that these associations are not explained solely by the between-method differences captured by the rank adjustment.

On the same texts and reference bank, increasing the horizon from 64 to 128 changes the variance-profile correlation by 0.139 ($[0.019,0.261]$). The associations remain sensitive to horizon, and ratings concern whole texts rather than isolated prefixes. Neither horizon establishes a diversity-DTW advantage over endpoint diversity. Appendix~\ref{app:length} provides extraction details, adjustment definitions, and paired comparisons.

\subsection{RQ3: Reference-Distribution Likelihood}
\label{sec:exploratory}
Reference likelihood also associates with ratings in the archived analysis (Table~\ref{tab:likelihood}). The mean-profile Gaussian has a correlation of 0.409 and the variance-profile Gaussian reaches 0.598. These findings are conditional on the cache reconstruction and extraction limitations described in Section~\ref{sec:features}.

Diagonal Gaussian and Gaussian-kernel estimates also yield positive associations (Table~\ref{tab:likelihood}). Domain-specific fits show further variation (Appendix~\ref{app:likelihood}). These comparisons do not identify a preferred distribution or domain on held-out data. The quantile plots in Figure~\ref{fig:qq} indicate departures from marginal normality; neither Gaussian nor Gaussian-kernel fits enforce the bounded support of diversity.

For RQ3, the comparisons suggest that the strength of the likelihood association varies with reference preparation and modeling choices. The positive variance-profile associations in the shared-pipeline experiment (Section~\ref{sec:length}) indicate that such an association also occurs with consistent extraction, although its magnitude depends on horizon. Their potential use in a new domain would need evaluation against that domain's texts and quality criteria. Figures~\ref{fig:controlled_density64} and~\ref{fig:controlled_density128} compare the controlled mean-profile distributions by decoding method with the fitted human reference. 

\section{Discussion and Practical Use}
The three facets offer different ways to relate generated text to human writing. Temporal alignment follows the evolution of the measured signals, summary comparison uses a human continuation of the same prompt, and likelihood places the joint profile within a broader human reference. The baseline comparisons assess how these different reference questions relate to quality prediction.

Across the experiments, diversity alignment and mean-based comparisons show positive associations with human judgments, while endpoint diversity also captures substantial variation in ratings. The paired comparisons leave a predictive advantage for DTW unresolved. An alignment may nevertheless be useful for examining where measured trajectories diverge. Establishing a connection between those differences and recognizable textual problems would require direct tests, for example with controlled repetition or topic shifts. Such perturbations can probe behavior beyond score--rating correlation \citep{he-etal-2023-blind}.

Reference likelihood adds a distributional perspective. The variance-profile associations in the controlled experiment suggest that typicality can carry quality-related information, although the strength of that relationship varies with the scoring horizon. An unusual profile may also reflect an appropriate but uncommon continuation. Applications such as filtering would therefore need evidence about which kinds of atypical text should be retained, as well as validation of the reference model in the intended domain.

Practical use also depends on the available references. Summary comparison needs a human continuation of the same prompt; temporal alignment and likelihood use human reference collections. Endpoint scores and fixed composites offer simpler alternatives when such resources are unavailable. Temporal alignment also has greater downstream computational cost than summary comparison or Gaussian scoring (Appendix~\ref{app:data}). 

The overall-quality judgments leave open which textual properties drive the associations. More targeted annotation could help connect the numerical patterns to readers' assessments of coherence and diversity.

\section{Conclusion}
We presented a reference-based framework for analyzing coherence and diversity in open-ended text generation through temporal alignment, comparison with matched human continuations, and likelihood under human reference distributions. The experiments suggest that diversity-based alignment and mean-based comparisons capture variation associated with human quality judgments. Positive associations also occur for reference likelihood, although their strength varies with the reference configuration and the portion of text evaluated. The comparisons with simpler baselines leave the predictive advantage of temporal alignment uncertain. The framework brings these findings together while making explicit the different questions that human references can support. Further work could examine whether the observed patterns correspond to recognizable textual properties, such as repetition or topic shifts, and whether these relationships extend to other generation settings.

\section*{Limitations}
\label{sec:limitations}
The evaluation covers a small English sample from one generator and fixed decoding configurations. Two annotators provide overall-quality ratings with limited agreement, leaving generalization and dimension-specific judgments open to further study. Further, Gaussian support extends beyond the feature bounds, and neither Gaussian nor KDE scores are calibrated quality probabilities. Bootstrap intervals condition on fixed references and PCA loadings; exploratory comparisons lack multiplicity adjustment. Strategy-adjusted associations remain descriptive rather than causal.

Finally, the measurements represent model predictability and lexical repetition, offering limited coverage of discourse coherence, creativity, factuality, or safety. Broader evaluation would require alternative assessor models, independent ratings, held-out settings, and controlled textual perturbations.

\section*{Acknowledgments}
Esteban Garc\'es Arias sincerely thanks the Mentoring Program of the Faculty of Mathematics, Statistics and Informatics at LMU Munich, and the Munich Center for Machine Learning (MCML) for their support. He also thanks the anonymous (meta) reviewers for their constructive feedback throughout the review process.

\section*{Ethics Statement}
We affirm that our research adheres to the \href{https://www.aclweb.org/portal/content/acl-code-ethics}{ACL Ethics Policy}. This work involves the use of publicly available datasets and does not include any personally identifiable information. An ethical concern worth mentioning is the use of language models for text generation, which may produce harmful content, either through intentional misuse by users or unintentionally due to the training data or algorithms. We declare that there are no conflicts of interest that could potentially influence the outcomes, interpretations, or conclusions of this research. All funding sources supporting this study are enumerated in the acknowledgments section. We have diligently documented our methodology, experiments, and results, and we commit to sharing our code, data, and other relevant resources to enhance reproducibility and further advancements in the field.

\bibliography{custom}

\clearpage
\appendix
\section*{Appendix}
\suppressfloats[t]
\section{Data and Reproducibility Details}
\label{app:data}
Figure~\ref{fig:trajectories} shows the first prompt in lexicographic identifier order (1017 Wikinews). The curves show the archived values without smoothing; their horizontal coordinate is the stored evaluation step.
Table~\ref{tab:data} summarizes ratings and stored lengths by method; Table~\ref{tab:reference_counts} documents the cached reference exclusions. The analysis preserves the original prompt and method identifiers. Machine continuations are matched to their human references by prompt identifier. Cached all-reference DTW rows are linked through their original observation identifiers and checked against both method and rating. Recomputed DTW comparisons use the same recurrence as the R \texttt{dtw} defaults. Recomputing all 25,000 machine--human comparisons (250 texts, 50 references, two signals) reproduces every cached average to numerical precision.

\begin{table}[t]
\centering\small
\setlength{\tabcolsep}{3pt}
\begin{tabular}{@{}lcc@{}}
\toprule
Method & Mean rating & Length range \\
\midrule
Human & 3.26 & 129--898 \\
Temperature & 2.42 & 20--631 \\
Top-$p$ & 2.69 & 11--266 \\
Top-$k$ & 2.54 & 48--408 \\
Beam & 2.00 & 115--270 \\
Contrastive & 2.30 & 22--483 \\
\bottomrule
\end{tabular}
\caption{Mean human ratings and observed trajectory lengths for each method, with 50 texts per method. Length counts the stored coherence measurements.}
\label{tab:data}
\end{table}

\begin{table}[t]
\centering\small
\begin{tabular}{@{}lrr@{}}\toprule
Domain & Complete & Retained \\
\midrule
BookCorpus & 1947 & 1947 \\
ROCStories & 2500 & 2500 \\
Wikinews & 2000 & 1976 \\
WikiText & 1314 & 1290 \\
WritingPrompts & 2164 & 2164 \\
\bottomrule\end{tabular}
\caption{Reference records with cached features before and after excluding exact overlaps with primary human continuations. The 75 missing-feature records are excluded before these counts.}
\label{tab:reference_counts}
\end{table}

For rating proximity, 82.7\% of pairs differ by at most one point and the mean absolute difference is 0.823. These supplement the agreement statistics in Section~\ref{sec:data}.

The extended collection contains 10,000 records: BookCorpus (1,947), ROCStories (2,500), Wikinews (2,000), WikiText (1,314), and WritingPrompts (2,239). BookCorpus, Wikinews, and WikiText pair presegmented leading text with its following passage. ROCStories pairs a title with the full story; WritingPrompts pairs a prompt with its response. We retain these boundaries. Recovering the original document-to-segment selection procedure is outside the available cache. Prefix length fields in the extended CSV count characters, not tokens.

Seventy-five WritingPrompts records lack cached metric arrays, leaving 9,925 complete records. Whitespace-normalized exact matching identifies 48 continuations also used as primary human references; excluding them leaves 9,877 records for archived reference fitting. The released analysis includes record counts and both exclusion masks. These checks do not rule out near duplicates or shared source documents.

\paragraph{Execution.}
The computational analysis uses Python, NumPy, pandas, SciPy, scikit-learn, and \texttt{dtw-python} 1.5.3. The analysis seed is 20260907. Plotting uses Matplotlib. OPT-2.7B inference supplies the controlled prefix experiment in Appendix~\ref{app:length}; the full-trajectory results use the existing caches. PCA uses the archived component scores, and MAUVE uses the archived corpus-level outputs. Source files and result tables accompany the manuscript.

\paragraph{Bootstrap implementation.}
We use 3,000 bootstrap replicates with seed 20260907. Each replicate samples prompt groups with replacement and retains all five machine outputs and their ratings within each group. The number of sampled groups equals the number evaluated: 50 for the full-trajectory analysis, or 40 and 23 for the fixed-horizon analyses. We refit each score's sample-dependent normalization within every replicate, including summary standardization and, for Q*Text, min--max ranges. The external reference corpus, temporal human-reference collection, and PCA loadings remain fixed. Intervals therefore describe uncertainty across the observed prompt groups conditional on these reference resources; they do not include uncertainty from sampling new annotators or refitting the reference corpus.

\paragraph{Computational costs.}
After feature extraction, summary computation is linear in trajectory length and a two-dimensional Gaussian score has constant evaluation cost. Standard DTW requires $O(|a||b|)$ work for each reference pair, multiplied here by 50 references; kernel density evaluation also scales with the reference collection. These downstream costs exclude language-model feature extraction, which may dominate. The summary facet additionally needs a human continuation of the evaluated prompt. A reference-free endpoint score or fixed composite can be preferable when those resources are unavailable.

\subsection{DTW Recurrence}
\label{app:dtw}
The temporal facet compares each machine trajectory $a$ with human trajectories $b$. We use DTW with scalar absolute local cost, closed endpoints, no warping window, and the \texttt{symmetric2} step pattern. With $\delta_{ij}=|a_i-b_j|$, the interior recurrence is
\begin{align}
G_{ij}=\min\{&G_{i-1,j}+\delta_{ij},\nonumber\\
&G_{i,j-1}+\delta_{ij},\nonumber\\
&G_{i-1,j-1}+2\delta_{ij}\},
\label{eq:dtw_recurrence}
\end{align}
with $G_{11}=\delta_{11}$ and boundary paths accumulating their local costs. The distance is $G_{|a|,|b|}$. Equation~\ref{eq:dtw_recurrence} fixes the step weights, which can affect the resulting scores.

Here $D(a,b)=G_{|a|,|b|}$ in Equation~\ref{eq:dtw}. The normalization uses the actual lengths of the two signal arrays.

\section{Additional Primary Results}
\label{app:primary}
\begin{table}[t]
\centering\small
\setlength{\tabcolsep}{3pt}
\begin{tabular}{@{}lcc@{}}
\toprule
Score & Rater A & Rater B \\
\midrule
Temporal: diversity & $0.507$ & $0.436$ \\
Temporal: coherence & $0.160$ & $0.090$ \\
Summary: mean & $0.554$ & $0.300$ \\
Summary: variance & $-0.521$ & $-0.363$ \\
Summary: all four & $-0.044$ & $-0.200$ \\
Endpoint diversity & $0.560$ & $0.276$ \\
Mean diversity & $0.543$ & $0.263$ \\
Variance diversity & $-0.573$ & $-0.290$ \\
Endpoint coherence & $-0.226$ & $-0.053$ \\
Fixed Q*Text & $0.599$ & $0.437$ \\
PCA: PC1 & $-0.423$ & $-0.204$ \\
PCA: PC2 & $-0.475$ & $-0.455$ \\
\bottomrule
\end{tabular}
\caption{Association of each score with the two annotators\textquotesingle{} individual quality ratings. Values are Spearman correlations on the same 250 machine texts.}
\label{tab:raters}
\end{table}

All correlations in Table~\ref{tab:raters} use the same 250 machine texts. The negative variance associations are reported with their original sign. For PCA, the orientation follows the released scores. A sign change in a principal component leaves its information content unchanged, so comparison of its raw signed correlation with a deliberately oriented quality score requires care.

The fixed Q*Text implementation normalizes each of the three endpoint inputs on the evaluation collection. For perplexity, normalization is $(\max p-p)/(\max p-\min p)$; for coherence and diversity it is $(v-\min v)/(\max v-\min v)$. Application to new texts requires specifying the normalization ranges or the collection used to estimate them. Its initialization is taken from the published Q*Text algorithm, rather than optimized on these ratings.

\paragraph{PCA.}
We use the released PCA scores from standardized coherence mean, coherence variance, diversity mean, and diversity variance. PCA was fitted without ratings on all 300 texts, including the human references; we evaluate its fixed components on the 250 machine texts. The first two components explain 58.23\% and 30.53\% of feature variance. Their signs are arbitrary, so a negative rating correlation does not by itself indicate a poor component. PCA maximizes feature variance, not quality prediction.

\paragraph{Fixed Q*Text.}
To include an untuned composite, we implement Q*Text's published initialization on the cached \emph{endpoint} perplexity, coherence, and diversity values \citep{arias2025betteropenendedtextgeneration}. After min--max normalization, with perplexity inverted, Equation~\ref{eq:qtext} defines the score:
\begin{equation}
Q_0=\frac{1}{3}\sum_{j=1}^{3}u_j\exp[-(u_j-0.5)^2].
\label{eq:qtext}
\end{equation}
Thus all weights and penalty strengths equal one, all targets equal 0.5, and no parameter is optimized against ratings. This specification defines the baseline evaluated here. The rating-tuned correlation of 0.555 reported by \citet{arias2025betteropenendedtextgeneration} uses a different evaluation protocol and is excluded from the common-sample comparison.

\paragraph{MAUVE.}
We retain the original notebook's corpus-level MAUVE results as descriptive context. Each comparison uses 50 human texts and 50 outputs from one strategy, GPT-2 features, and maximum text length 512. The notebook supplies the complete stored text field, including prompt and continuation labels. This evaluates distributions of those text fields, not a per-continuation quality score. These comparisons yield five corpus-level scores, one per decoding strategy.

\section{Reconstructed Reference Distributions}
\label{app:likelihood}
\begin{figure*}[t]
\centering\includegraphics[width=\textwidth]{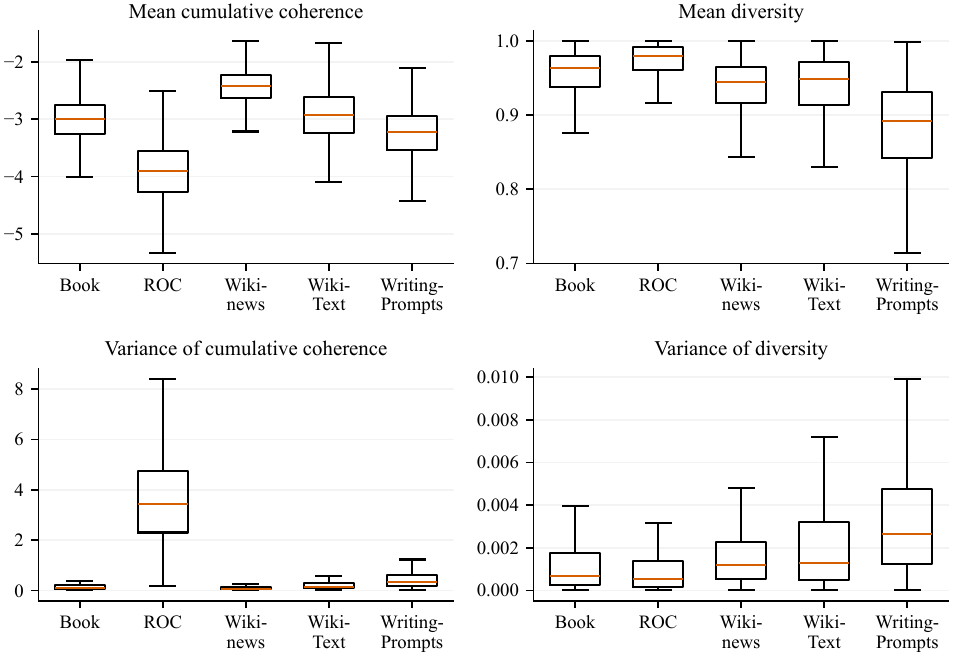}
\caption{Coherence and diversity summaries across human reference domains. Boxes show quartiles and 1.5-IQR whiskers for the reconstructed, overlap-excluded records. Outliers are hidden for readability but retained in fitting. Each panel uses a scale suited to its measured property.}
\label{fig:domains}
\end{figure*}

\begin{figure*}[t]
\centering\includegraphics[width=\textwidth]{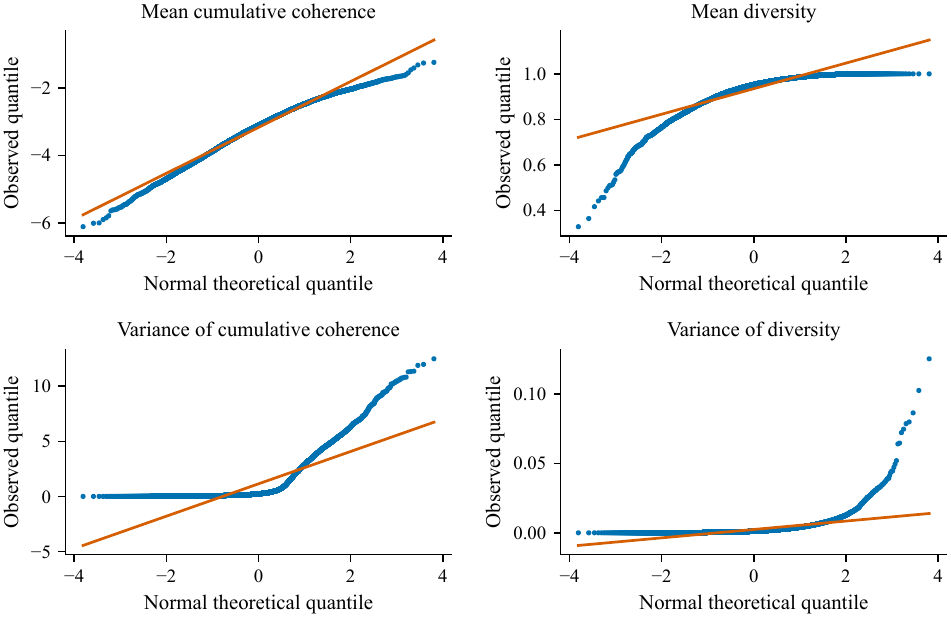}
\caption{Comparison of the four human reference summaries with normal distributions. Departures from the straight reference lines indicate differences from marginal normality. All fitted records are shown; these marginal checks complement the joint distributions in Figure~\ref{fig:reference}.}
\label{fig:qq}
\end{figure*}

\begin{figure*}[t]
\centering\includegraphics[width=\textwidth]{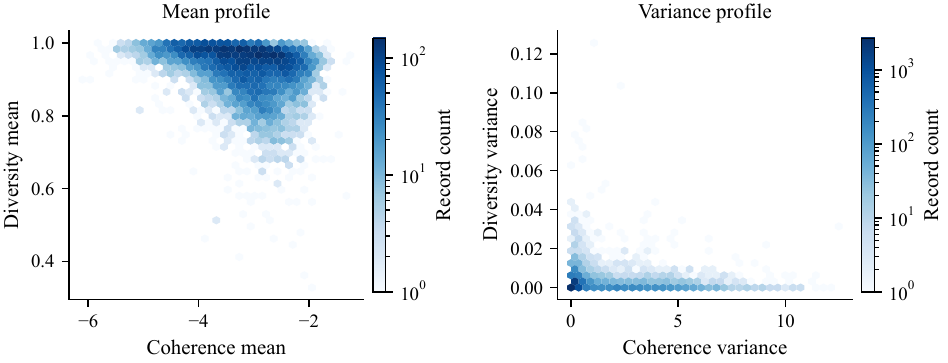}
\caption{How coherence and diversity summaries vary together in the reconstructed human reference corpus: means (left) and variances (right). Darker hexagons contain more records, on the logarithmic count scales shown. The panels visualize the observations used to fit the reference models.}
\label{fig:reference}
\end{figure*}

The fitted parameters below refer to 9,877 complete reference records after exact-overlap removal. Figure~\ref{fig:domains} shows how the reconstructed summaries vary across reference domains, Figure~\ref{fig:qq} compares their marginal distributions with normal quantiles, and Figure~\ref{fig:reference} shows their joint distributions. These views help interpret the model and domain sensitivity checks in Section~\ref{sec:exploratory}. Per-record coherence is first converted from stored token log probabilities to cumulative averages. Per-text variances use divisor $n$, while the covariance of the resulting reference vectors uses divisor $N-1$.
Fitting each reference domain separately gives another view of reference sensitivity: mean-profile correlations range from 0.188 (WritingPrompts) to 0.569 (Wikinews), while variance-profile correlations range from 0.512 (ROCStories) to 0.595 (WritingPrompts). All fits use the same machine evaluation texts, so these ranges describe sensitivity to the reference domain within this sample.

\begin{table}[t]
\centering\small
\setlength{\tabcolsep}{3pt}
\begin{tabular}{@{}lcc@{}}
\toprule
Score & $\rho$ & 95\% interval \\
\midrule
Gaussian: mean & $0.409$ & $[0.315,0.493]$ \\
Diagonal Gaussian: mean & $0.441$ & $[0.346,0.523]$ \\
KDE: mean & $0.436$ & $[0.332,0.522]$ \\
Gaussian: variance & $0.598$ & $[0.505,0.675]$ \\
Diagonal Gaussian: var. & $0.604$ & $[0.511,0.679]$ \\
KDE: variance & $0.624$ & $[0.520,0.707]$ \\
\bottomrule
\end{tabular}
\caption{Association between reconstructed reference likelihood scores and human ratings on 250 machine texts. Values are Spearman correlations with 95\% prompt-cluster bootstrap intervals. The archived extraction limitations are described in Appendix~\ref{app:audit}.}
\label{tab:likelihood}
\end{table}

\begin{align*}
\mu_{\mathrm{mean}}&=(-3.173622,\ 0.934842)^T,\\
\Sigma_{\mathrm{mean}}&=\begin{bmatrix}0.4668553&-0.0123085\\-0.0123085&0.0038644\end{bmatrix}.
\end{align*}
\begin{align*}
\mu_{\mathrm{var}}&=(1.147170,\ 0.002390)^T,\\
\Sigma_{\mathrm{var}}&=\begin{bmatrix}3.2443632&-0.0005244\\-0.0005244&0.0000207\end{bmatrix}.
\end{align*}


\section{Feature Preparation and Sensitivity}
\label{app:audit}
For each primary text, the stored coherence array has $L$ values and the diversity array has $L-1$. All per-text means and population variances used below are recomputed over the actual stored arrays; they match the released summary columns to numerical precision. The coherence length $L$ is the denominator in the summary normalization. DTW uses each signal's own array length. We do not pad one array to match another.

The extended cache requires separate preparation. Its coherence arrays contain token log probabilities, and its saved summary values are rounded to two decimal places. We form cumulative averages of the stored log probabilities before calculating trajectory means and population variances, and recompute diversity summaries from the stored arrays. This avoids treating rounded zero variances as evidence of constant trajectories. The underlying arrays remain rounded, and the primary and reference extraction pipelines are not fully documented. We therefore interpret the reconstructed likelihood results as \emph{exploratory} evidence about the association between reference typicality and ratings. The sensitivity comparisons below assess the consequences for the fitted scores.

As an exploratory alternative, we calculate the negative Euclidean distance from a text's two-dimensional summary to the reconstructed corpus mean. Unlike the prompt-relative composite, this alternative changes the reference, coordinate scaling, and aggregation together. Its performance cannot isolate the benefit of prompt-specific references.

The extended summary CSV contains two-decimal summaries. In particular, small nonzero diversity variances may be stored as zero. The supplied variance-reference parameters were fitted to 1,156 records with positive stored diversity variance. Such selection changes the modeled population: it is a distribution conditional on crossing a rounding threshold, rather than a reference for all human records. The reconstruction calculates variances from the available arrays and retains genuine zero-variance trajectories.

On the common 250-machine subset, applying the supplied summary-based Gaussian parameters gives correlations of 0.458 for means and 0.257 for variances. The reconstructed, overlap-excluded fits give 0.409 and 0.598. This comparison describes sensitivity to data preparation on the same evaluation sample. The summary-based and array-based configurations differ in multiple ways, so the comparison does not isolate a single cause.

All primary temporal coefficients are recomputed as Spearman correlations. On the same 250-machine subset, the negative Euclidean mean-profile score has $\rho=0.165$ and its variance counterpart has $\rho=0.019$. These scores use the reconstructed corpus means, retain the raw coordinate scales, and change several design choices relative to the prompt-relative composite. Their weaker observed associations do not isolate reference selection as the cause. These cached continuation-level comparisons use the same 250 machines; the controlled prefix experiment uses the eligible subsets in Appendix~\ref{app:length}. Corpus-level MAUVE remains a separate descriptive analysis.

\section{Length Controls and Horizon Sensitivity}
\label{app:length}
\paragraph{Shared extraction.}
We score all 300 primary and 10,000 reference texts using OPT-2.7B, model/tokenizer revision \texttt{905a4b6}, in float32 with TF32 disabled. Prompt and continuation whitespace is stripped separately; each is tokenized separately without special tokens, and exactly one BOS token is prepended. We score at most the first 128 continuation tokens, conditioned on the prompt. All prompts have at most 81 tokens, so these measurements fit within the 2,048-token context. Coherence is the cumulative mean of next-token log probabilities. Diversity is the product of distinct-to-total token-ID $n$-gram ratios for $n=2,3,4$, with ratios set to one before enough tokens exist. Both signals include the first continuation token and use identical positions. Per-text variances use divisor $T$. Validation compared token IDs, lengths, and hashes with the source texts and recomputed summary features from the stored trajectories. Cross-entropy and block-partition checks verified the consistency of the GPU calculations. Full model and tokenizer identifiers, software versions, and validation results accompany the analysis.

\paragraph{Eligibility and references.}
Horizons 64 and 128 were selected from cached length coverage before the controlled measurements, without optimizing against ratings; this is not a preregistered design. Table~\ref{tab:controlled} summarizes the resulting associations. Only complete six-text prompt groups with at least $T$ tokens in every continuation are retained. This gives 40 groups (200 machines) at 64 and 23 groups (115 machines) at 128. No sequence is padded. The DTW human bank contains the corresponding 40 or 23 humans; distances use symmetric2 absolute-cost alignment divided by $T$. Gaussian fits use 7,686 and 7,149 eligible human-corpus records after excluding 48 exact overlaps with primary human references. Both reference and evaluated summaries use the same horizon and extraction. The controlled extraction includes the 75 texts without archived feature arrays. Eligibility alters domain composition: no ROCStories records qualify at 128.

\paragraph{Inference and paired horizon comparison.}
Intervals use 3,000 prompt-cluster bootstrap replicates (seed 20260907), retaining all five machine methods per sampled prompt and conditioning on fixed reference banks. Summary standardization is refitted in each replicate. For rank-residual sensitivity, both score and rating ranks are residualized against an intercept and cubic polynomials of centered empirical ranks of original machine continuation length, matched-human continuation length, and prompt length. Ranks and regressions are refitted per replicate. These are descriptive residual associations, not ordinary Spearman coefficients or causal effects. The adjusted diversity-DTW and Gaussian variance-profile coefficients are 0.336 and 0.286 at 64 tokens, and 0.439 and 0.493 at 128.

To isolate horizon changes from changes in cohort and reference membership, we additionally score 64-token prefixes on the 128-eligible set: the same 115 machines, 23 primary humans, and 7,149 corpus references. Gaussian parameters are refitted at each horizon on those same records. The paired 128-minus-64 changes are 0.121 ($[-0.036,0.290]$) for diversity DTW, 0.066 ($[-0.008,0.137]$) for Gaussian means, and 0.139 ($[0.019,0.261]$) for Gaussian variances. These intervals are not multiplicity-adjusted. The paired differences between diversity DTW and endpoint diversity are $-0.058$ ($[-0.201,0.086]$) at 64 and $-0.107$ ($[-0.270,0.062]$) at 128; neither establishes superiority or equivalence.

\begin{table*}[t]
\centering\small
\begin{tabular}{lcc}
\toprule
Score & 64 tokens & 128 tokens\\
\midrule
DTW coherence & -0.145 [-0.316, 0.043] & 0.031 [-0.179, 0.244] \\
DTW diversity & 0.275 [0.164, 0.385] & 0.362 [0.147, 0.529] \\
Gaussian mean & -0.024 [-0.198, 0.148] & 0.180 [0.037, 0.310] \\
Gaussian variance & 0.298 [0.137, 0.435] & 0.531 [0.337, 0.675] \\
Endpoint diversity & 0.342 [0.207, 0.468] & 0.501 [0.350, 0.623] \\
Summary mean & 0.317 [0.122, 0.476] & 0.370 [0.158, 0.549] \\
\bottomrule
\end{tabular}
\caption{Associations with human ratings after adjusting score and rating ranks for decoding strategy. Values are correlations of the residual ranks with 95\% prompt-cluster bootstrap intervals. Texts and scores match Table~\ref{tab:controlled}; references remain fixed during resampling.}
\label{tab:strategy}
\end{table*}

\paragraph{Decoding-strategy sensitivity.}
Table~\ref{tab:strategy} reports an exploratory adjustment applied to all six scores in Table~\ref{tab:controlled}. We rank scores and ratings over each eligible cohort using average ranks for ties, regress each rank vector on five decoding-method indicators, and correlate the residuals. This removes method-specific mean ranks; it is neither an average of within-method Spearman correlations nor a causal adjustment. Each of 3,000 prompt-cluster bootstrap replicates (seed 20260908) refits summary standardization, ranks, and regressions, with reference banks fixed. Intervals are pointwise and not multiplicity-adjusted. This analysis controls decoding strategy separately from the length adjustment above, rather than estimating a joint adjustment. Diversity DTW, variance-profile likelihood, endpoint diversity, and the mean summary retain positive residual associations at both horizons. Coherence DTW has intervals spanning zero at both horizons, and mean-profile likelihood at 64 tokens also has an interval spanning zero. The adjustment thus leaves different patterns of association across the measured signals and facets.

\paragraph{Interpretation.}
Equal token horizons hold the number of scored continuation tokens constant within each analysis. Content differences and other length-related influences may remain, and the results can vary with the horizon. The ratings describe entire continuations, so prefix scores are not validated against prefix-specific judgments. The prefix-only run does not provide full-length scores beyond 128 tokens. Eligibility restricts the texts represented in these results. The Gaussian scores describe typicality under the fitted model, with their relationship to quality evaluated through correlation.

\begin{figure*}[t]
\centering\includegraphics[width=\textwidth]{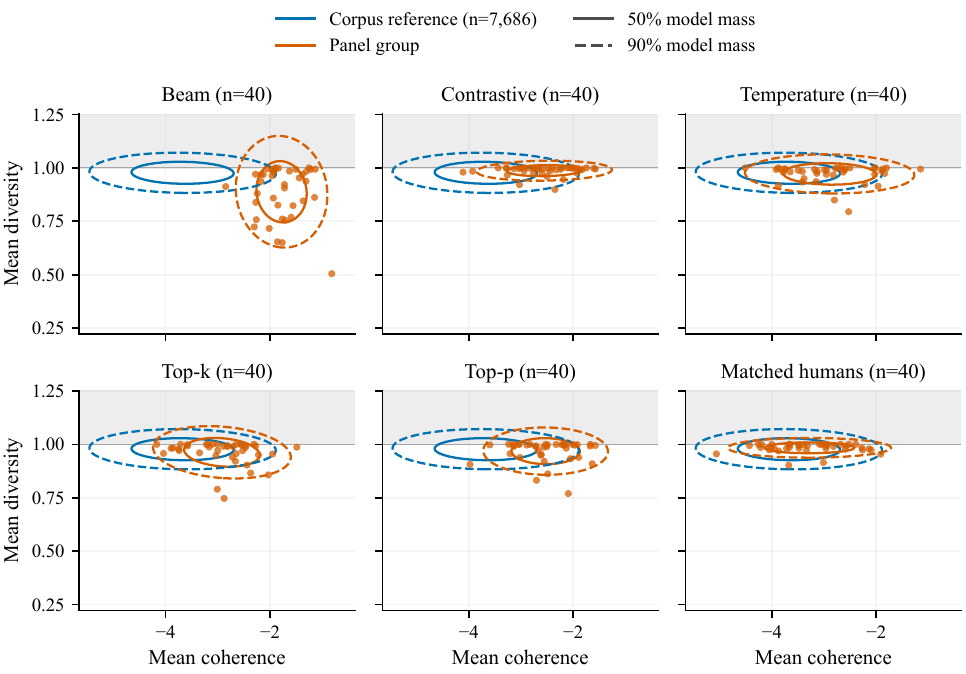}
\caption{Human and machine coherence--diversity profiles at 64 continuation tokens. Each panel shows 40 continuations from one group. Blue contours describe the human corpus reference (7,686 records); orange contours describe the panel group. Solid and dashed curves enclose 50\% and 90\% of fitted Gaussian mass, rather than confidence intervals. Gray shading marks values beyond the diversity range. Axes match Figure~\ref{fig:controlled_density128}.}
\label{fig:controlled_density64}
\end{figure*}

\begin{figure*}[t]
\centering\includegraphics[width=\textwidth]{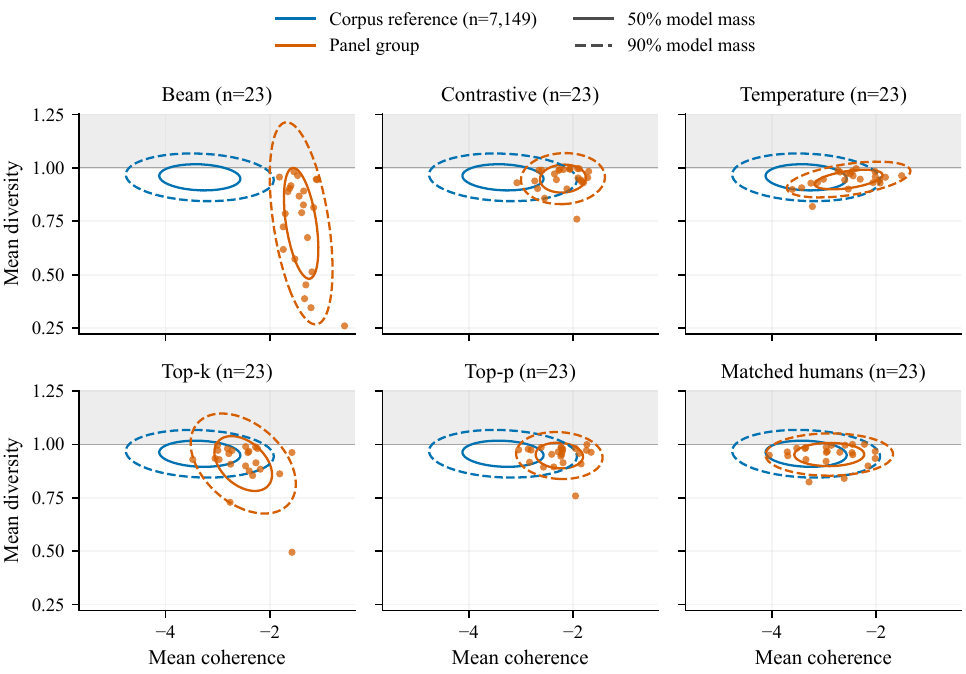}
\caption{Human and machine coherence--diversity profiles at 128 continuation tokens. Each panel shows 23 continuations. Blue contours describe the human corpus reference (7,149 records); orange contours describe the panel group. Contour masses, shading, and axes match Figure~\ref{fig:controlled_density64}. Only the corpus reference supplies the likelihood scores in Table~\ref{tab:controlled}.}
\label{fig:controlled_density128}
\end{figure*}

\section{Controlled Human--Machine Density Comparisons}
\label{app:density_comparison}
Figures~\ref{fig:controlled_density64} and~\ref{fig:controlled_density128} compare human and machine profiles using the controlled measurements described in Appendix~\ref{app:length}. Each point is a continuation's mean cumulative coherence and mean token-level diversity at the indicated horizon. The five machine panels and matched-human panel use the same complete prompt groups within each horizon. Both figures have identical axes, and every eligible primary observation is included without jitter or selection by rating. Overlapping points may obscure individual observations.

Blue contours use exactly the corpus-reference mean and covariance used by the controlled mean-profile likelihood score. Orange contours use a separate Gaussian fitted to the observations in that panel, with sample covariance. Panel-specific fits summarize each group's observations; likelihood scores use the human corpus reference. Solid and dashed curves enclose 50\% and 90\% of their respective fitted Gaussian distributions. They are neither confidence intervals nor estimates of empirical coverage. The curves satisfy $(u-\mu)^T\Sigma^{-1}(u-\mu)=\chi^2_2(q)$ for mass $q$, and were checked numerically against this identity.

The beam panel illustrates that group differences can involve both location and spread. At 128 tokens, its observed mean-diversity values span a visibly broader range than those of the matched humans. This is descriptive evidence about these measured profiles, not a test that one strategy produces better text. The gray region above diversity one makes a limitation of the Gaussian approximation explicit: its contours can extend beyond the signal's possible range. The mean-profile views do not establish the fit or validity of the separate variance-profile likelihood model.

The 64-token figure uses 40 texts per panel and 7,686 corpus-reference records. The 128-token figure uses 23 texts per panel and 7,149 references. These are the cohorts in Appendix~\ref{app:length}; differences between the figures combine horizon and eligibility changes. The fixed-membership comparison in that appendix examines horizon differences while holding the evaluated texts and references constant.

\FloatBarrier

\end{document}